\documentclass[10pt,twocolumn,letterpaper]{article}

\usepackage[pagenumbers]{cvpr} 

\usepackage{graphicx}
\usepackage{amsmath}
\usepackage{amssymb}
\usepackage{booktabs}

\usepackage[pagebackref,breaklinks,colorlinks]{hyperref}

\usepackage[capitalize]{cleveref}
\crefname{section}{Sec.}{Secs.}
\Crefname{section}{Section}{Sections}
\Crefname{table}{Table}{Tables}
\crefname{table}{Tab.}{Tabs.}

\usepackage{placeins}
\usepackage[ruled,vlined]{algorithm2e}
\usepackage{array,multirow}

\def\confName{AI4CC Workshop at CVPR}
\def\confYear{2022}

\begin{document}

\title{SaiNet: Stereo aware inpainting behind objects with generative networks}

\author{Violeta Menéndez González\textsuperscript{1,2}\\
{\tt\small v.menendezgonzalez@surrey.ac.uk}
\and
Andrew Gilbert\textsuperscript{1}\\
{\tt\small a.gilbert@surrey.ac.uk}
\and
Graeme Phillipson\textsuperscript{2}\\
{\tt\small graeme.phillipson@bbc.co.uk}
\and
Stephen Jolly\textsuperscript{2}\\
{\tt\small stephen.jolly@bbc.co.uk}
\and
Simon Hadfield\textsuperscript{1}\\
{\tt\small s.hadfield@surrey.ac.uk}
\and
\textsuperscript{1}CVSSP, University of Surrey \quad \textsuperscript{2}BBC R\&D
}

\twocolumn[{%
  \renewcommand\twocolumn[1][]{#1}%
\maketitle
\thispagestyle{empty}
\begin{center}
  \newcommand{\teaserwidth}{0.9\textwidth}
  \vspace{-0.2in}
  \centerline{
    \includegraphics[width=0.8\linewidth]{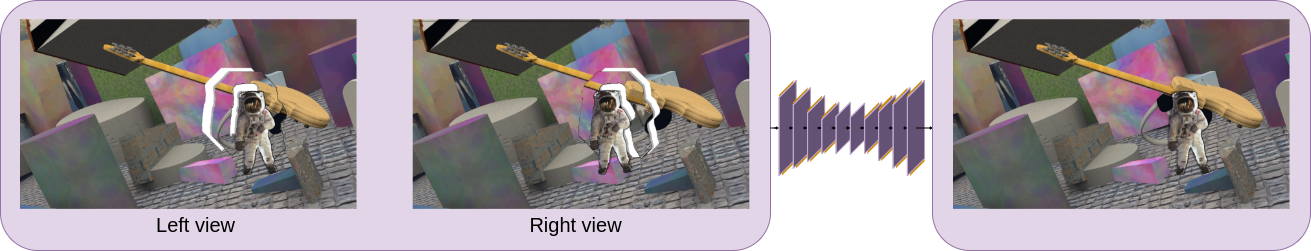}
    }
    \captionof{figure}{\textbf{SaiNet:} Inpainting behind objects using geometrically meaningful masks.}
  \vspace{-0.1in}
  \label{fig:teaser}
 \end{center}%
}]

\begin{abstract}
   In this work, we present an end-to-end network for stereo-consistent image inpainting with the objective of inpainting large missing regions behind objects. The proposed model consists of an edge-guided UNet-like network using Partial Convolutions. We enforce multi-view stereo consistency by introducing a disparity loss. More importantly, we develop a training scheme where the model is learned from realistic stereo masks representing object occlusions, instead of the more common random masks. The technique is trained in a supervised way. Our evaluation shows competitive results compared to previous state-of-the-art techniques.
\end{abstract}

\section{Introduction}
\label{sec:intro}

Image inpainting is the task of filling in the missing regions of an image with perceptually plausible content. It has many vital applications in computer vision and image processing: the removal of unwanted objects (e.g. superimposed text), image and film restoration (e.g. scratches or cracks), image completion (e.g. dis-occlusions), cinema post-production, among others. Our work focuses on the under-explored problem of stereo-inpainting. This lends itself to applications that need to see behind objects to generate reasonable image fillers, for example in novel view synthesis of scenes, object removal in stereoscopic video, 3D animation of still images, and dis-occlusion in virtual reality environments.

This paper focuses on applications of inpainting which may improve view synthesis in media production, this requires an approach that can take advantage of multiple cameras, but doesn't necessarily have the computer capacity of other novel view approaches that reconstruct a whole 3D scene representation. In addition, we want an approach that can generalise well to unseen scenes and generate creative content without human input, therefore we want to apply CNNs that can single-handedly propagate structures and textures reasonably.

In previous works, traditional monocular techniques tried to achieve inpainting by propagating local image structures and textures or copying patches from the known areas of the image. This worked well for small or narrow regions, but it was prone to generating visual inconsistencies in more significant gaps. Early stereo techniques attempted to equivalently generate consistent image output by mechanically warping the available data from the other views~\cite{wang2008}, or completing the disparity images~\cite{morse2012, mu2014}, and then proceeding similarly to the monocular inpainting approaches. However, in recent years, Deep Learning (DL) techniques have taken advantage of large-scale training data to create more semantically significant inpainting outputs. Some works focused on learning embeddings of the images~\cite{pathak2016,iizuka2017}, while others developed different types of convolutional layers to be able to handle more realistic irregular holes~\cite{liu2018,yu2019a}. However, the only DL techniques that address the stereo inpainting problem~\cite{chen2019,ma2020, luo2020} to date have focused on artificial or unrealistic inpainting regions, or don't enforce multi-camera consistency.

In contrast, our approach focuses on inpainting one target image on geometrically meaningful masks while using the information available from the other viewpoint. Our network architecture is inspired by the 3D photography generation work of Shih \etal~\cite{Shih2020} using a Partial Convolution~\cite{liu2018} architecture, which optimises the use of irregular masks at random locations. Furthermore, we improve the inpainting task by adding colour edge information following the idea by Nazeri \etal~\cite{Nazeri2019} in their work with EdgeConnect.

More importantly, we propose a novel stereo inpainting training mechanism. Instead of using random image masks, which usually represent the physical damage a picture can suffer, we use meaningful and geometrically-consistent object masks that are not necessarily bounded within the image. We extend the 2D context/synthesis region approach proposed by \cite{Shih2020} to use a bank of geometrically-consistent 3D object masks. Ground-truth training examples are generated from random virtual 3D objects placed at random locations in the foreground of the scene, allowing us to have a fully self-supervised stereo training approach. This data augmentation process addresses both the significance of masked regions and the stereo data scarcity problem. Furthermore, the resulting model is computationally efficient and able to generalise to previously unknown scenes and occluding objects.

In summary, the contributions of this paper are:
\begin{itemize}
    \item A novel stereo-aware structure-guided inpainting model suitable for efficient novel-view synthesis and free viewpoint VR applications.
    \item First inpainting work to take full advantage of stereo-context with geometrically-consistent object masks.
    \item A novel stereo consistency loss attempting to ensure that inpainting results are consistent with disoccluded information present in other views.
\end{itemize}

\section{Background}
\label{sec:background}

\paragraph{Learnable inpainting}

With the advancements of Deep Learning and the availability of large-scale training data, deep Convolutional Neural Networks (CNNs) became a popular tool for image prediction. Initial CNN models attempted to perform image inpainting by using feature learning with \textit{Denoising Autoencoders}~\cite{xie2012}, translation variant interpolation~\cite{ren2015}, or exploiting the shape of the masks~\cite{kohler2014}. Yet all these methods were only applicable to tiny and thin masks and lacked semantic understanding of the scenes. With the addition of \textit{Generative Adversarial Networks} (GANs)~\cite{goodfellow2014}, CNN architectures were able to extract meaningful semantics from images and generate novel content. Pathak \etal~\cite{pathak2016} used an encoder-decoder architecture to create a latent feature representation of the image, which captured both the semantics and appearance of structures, but struggled to maintain global consistency. Iizuka \etal~\cite{iizuka2017} proposed using both local and global context discriminators, which helped the local consistency of generated patches and still held image coherence in the whole. Yu \etal~\cite{yu2018a} added a contextual attention layer to aid the modelling of long-term correlations between the distant information and the hole regions.

Traditional vanilla convolutions depend on the hole initialisation values, which usually leads to visual artefacts. Liu \etal~\cite{liu2018} proposed the use of \textit{Partial Convolutions}: masked and re-normalised convolutional filters conditioned only on valid pixels. Yu \etal~\cite{yu2019a} extended this idea with \textit{Gated Convolutions} by generalising to a learnable dynamic features selection mechanism. Previous works focused on centred rectangular holes, which may cause methods to overfit to this kind of mask. Masked convolutions allowed models to handle more realistic irregular holes. Liu \etal~\cite{liu2018} studied the effects when the holes are in contact with the image border and created a large benchmark of irregular masks with varying sizes and locations. Many of these methods still fail to reconstruct reasonable structures and usually over-smooth surfaces. Some approaches \cite{song2018,Nazeri2019,Ren2019} tackle this problem by trying first to recover structural information to guide the inpainting of fine details and textures. With a two-stage adversarial model, \textit{EdgeConnect}~\cite{Nazeri2019} first recovers colour edges, while \textit{StructureFlow}~\cite{Ren2019} choose edge-preserved smooth images as the global structure information.
\paragraph{Stereo Consistent Inpainting}
There is little research done on stereoscopic image inpainting in the framework of deep learning. Following a similar trajectory to monocular approaches, traditional patch-based methods \cite{morse2012,mu2014,wang2008} find example patches from the available parts of the image and fill the holes applying consistency constraints. Wang \etal~\cite{wang2008} simultaneously inpaint colour and depth images using a greedy segmentation-based approach, inpainting first partial occlusions using warping, and total occlusions with a depth-assisted texture synthesis technique. Morse \etal~\cite{morse2012} extend \textit{PatchMatch}~\cite{barnes2009} to cross-image searching and matching without explicit warping, using a completed disparity map to guide the colour inpainting. Multi-view inpainting techniques such as Gilbert \etal~\cite{gilbert2018} create a dictionary of patches from multiple available viewpoints that are then coherently selected and combined to recover the missing region.

\begin{figure*}[!htbp]
\begin{center}
    \includegraphics[width=1\linewidth]{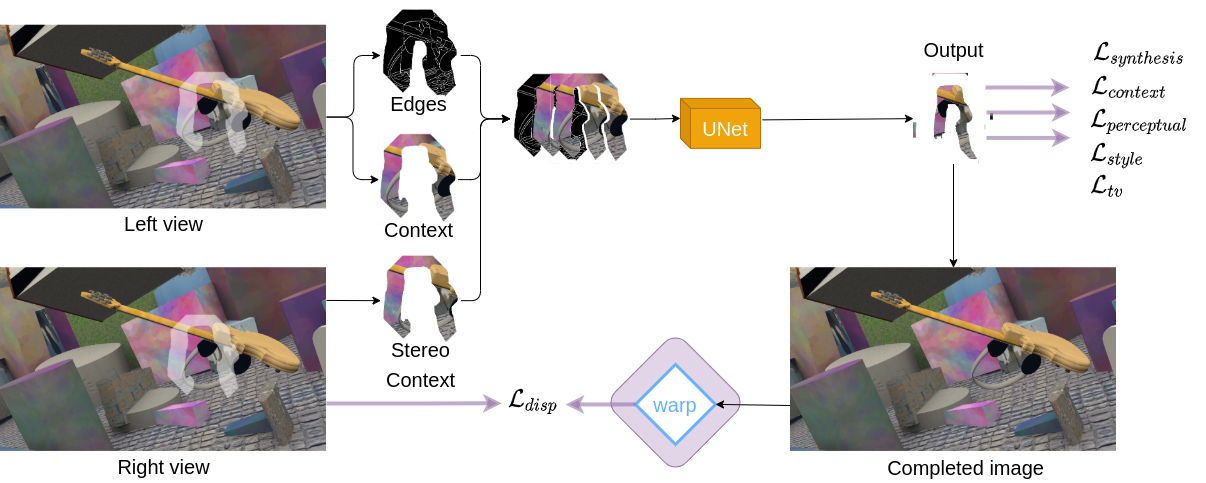}
\end{center}
   \caption{\textbf{Model overview:} Edge guidance, stereo context and disparity loss.}
\label{fig:model_overview}
\end{figure*}

The first stereo inpainting approach using deep learning was made by Luo \etal~\cite{luo2020}. They use a double reprojection technique to generate image occlusion masks from several new viewpoints, then apply \textit{Partial Convolutions}~\cite{liu2018} to inpaint the holes, and aggregate the results in a layered depth image. This technique shows good visual results on their own Keystone B\&W dataset, but they don't take into account multi-view consistency and rely on depth maps to reconstruct the image. Other techniques take advantage of both left and right views, like Chen \etal~\cite{chen2019}. They use an extension of \textit{Context Encoder}~\cite{pathak2016}. They inpaint left and right views simultaneously encoding both views and aggregating them at the feature level. In addition, they introduced a local consistency loss which helps preserve the inpainting consistency at a pixel level. They applied this model to inpainting regular holes at the centre of the image. Ma \etal~\cite{ma2020} use a similar architecture for two different tasks: reconstructing missing objects in one view that are available on the other view and coherently inpainting the same holes in both views similar to~\cite{chen2019}. To do this, they use two different stereo consistency losses, a warping-based consistency loss and a stereo-matching PSMNet-based~\cite{chang2018} disparity-reconstruction loss. However, because of a lack of ground-truth data for object removal, they only train their model on corruption restoration data. In contrast, our approach uses realistic and geometrically consistent foreground object masks to explore inpainting behind objects in stereo scenes.

\section{Approach}
\label{sec:approach}

\subsection{Model overview}

An overall visualisation of our proposed model can be seen in Figure~\ref{fig:model_overview}. It consists of a deep neural network that follows a UNet-like architecture~\cite{ronneberger2015} with partial convolutions~\cite{liu2018}. The network takes the context and synthesis areas of an object, where the context area is the background surrounding the object. The synthesis area is the region behind the object (hole) that the network will inpaint. In addition, the colour edges are fed into the network for structural guidance. Finally, to enrich the inpainting and make the network aware of the stereo view, a stereo-context image is added to the input.

\subsection{Object occlusion regions}\label{subsec:object_background_regions}
An essential part of image inpainting specifies the type of missing regions that the model needs to handle. Most previous approaches to inpainting have focused on randomly shaped inpainting masks of limited complexity. This is reasonable when dealing with image degradations such as scratches or removing regions containing nuisance objects in 2D. However, for stereo inpainting where we wish to maintain crisp object boundaries, this approach no longer makes sense. We need inpainting masks that represent real image occlusions. Therefore we propose a self-supervised approach where stereoscopic scenes are augmented with geometrically valid inpainting masks, based on a virtual 3D occluding object. This object is hallucinated in a stereo-consistent way over both images, which allows us to collect \textit{``behind the object''} ground-truth data. As such, the network learns to fill in geometrically-meaningful holes with background information, which can then be applied to actual object occlusions in novel view synthesis applications.

\begin{figure*}[htbp]
	\begin{center}
        \includegraphics[width=0.7\linewidth]{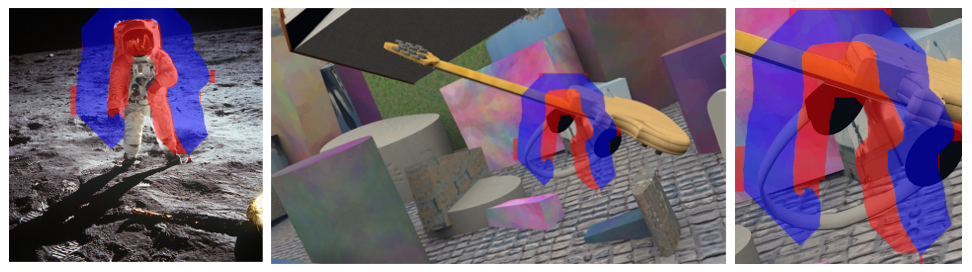}
    \end{center}
	
	\caption{\textbf{Data generation process.} A collection of \textcolor{blue}{context}/\textcolor{red}{synthesis} regions is created by extracting them from object boundaries in images on the COCO dataset. Then they are randomly sampled, warped, and pasted onto different images, forming the training dataset of ground truth context/synthesis regions}
	\label{fig:data-generation}
\end{figure*}

We generate these geometrically-valid masks from an unrelated dataset of natural scenes with either ground truth depth or object segmentations. As summarised in Algorithm \ref{alg:mask-generation}, we first detect depth discontinuities~\cite{Shih2020} along object boundaries and generate context and synthesis regions by propagating this boundary towards the background image (context), and the foreground object area (synthesis) (See Fig. \ref{fig:data-generation} for a visualisation of these areas). This way, we create a geometrically-meaningful bank of masks for inpainting.

\begin{algorithm}[b]
\SetAlgoLined
\KwIn{$\mathcal{N} = \{\mathbf{I}: \mathbf{I} \text{ is a natural image}\}$}
\KwOut{$\mathcal{M} = \{(\mathbf{C}^{obj}, \mathbf{S}^{obj}) \mid \forall obj \in \mathbf{I}, \forall \mathbf{I} \in \mathcal{N}\}$}
 \For{$\mathbf{I}$ in $\mathcal{N}$}{
 Find set of discontinuities $d_\mathbf{I} \equiv \{d^{obj}_\mathbf{I} \mid obj \text{ is an object in image } \mathbf{I}\}$\;
      \For{$d^{obj}_\mathbf{I}$ in $d_\mathbf{I}$}{
      Propagate background around $d^{obj}_\mathbf{I}$ to generate context mask $\mathbf{C}^{obj}$\;
      Propagate foreground around $d^{obj}_\mathbf{I}$ to generate synthesis mask $\mathbf{S}^{obj}$\;
      }
 }
 \caption{Generation of geometrically-valid masks}
 \label{alg:mask-generation}
\end{algorithm}

These masks are varied and irregular, preventing our model overfitting one type of mask. Furthermore, as opposed to most methods, our model doesn't use the whole image as context for the inpainting process, but just the region closer to the object boundaries. Although this reduces the available information that the network can learn from, it allows the network to narrow its attention to the most relevant and meaningful area. However, this poses a more challenging problem, as the context-to-synthesis area ratio is smaller, and the masked regions are not necessarily bounded by context on all sides.

\subsection{Stereo awareness}

Our approach aims to make inpainting consistent across views in two different ways. One is by enriching the network with extra available information. The other is by enforcing a disparity loss on the output of the network (as explained in Section \ref{subsec:disp_loss}). 
The main advantage of having two (or more) cameras is the additional information we can extract to make the task of inpainting ``unknown'' areas easier. For example, some colours or textures may be completely occluded by the object in one view, but still be partially visible from the other view. In this case, the additional input can provide strong cues for the network to inpaint the occluded region. We make our system stereo-aware by providing this extra information as input to the network by warping the context mask of each object based on its estimated disparity value into the additionally available view (See Algorithm~\ref{alg:data-generation}). We use \textit{PSMNet}~\cite{chang2018} to estimate this disparity and select the closest depth value to make sure the object is situated at the front of all other objects in the scene. In other words, we extract contextual information around the boundary of the occluding object, in both views. Then we feed this extra context into the network to learn to use it in filling in the synthesis area. In this way, we aid the inpainting process by enriching the texture and colour information available.

\begin{algorithm*}[htb]
\SetAlgoLined
\KwIn{$\mathcal{D} = \{(\mathbf{I}_L,\mathbf{I}_R): \text{stereo pair of images}\}$\\
 $\mathcal{M} = \{(\mathbf{C}^{obj}, \mathbf{S}^{obj}) \mid \forall \text{ object } obj\}$}
\KwOut{$\mathrm{Training\_set} = \{(\mathbf{CC}_L,\mathbf{CS}_L,\mathbf{E}_L,\mathbf{CC}_R) \mid \forall (\mathbf{I}_L,\mathbf{I}_R) \in \mathcal{D}\}$}
 \For{$(\mathbf{I}_L,\mathbf{I}_R)$ in $\mathcal{D}$}{
  1. Select random context and synthesis masks $\mathbf{C}^{obj}, \mathbf{S}^{obj}$ from the $\mathrm{Mask\_Bank}$\;
  2. Select a random position $x, y$ to situate the object at $\mathbf{I}_L$\;
  3. Crop image $\mathbf{I}_L$ at $x, y$ with mask $\mathbf{C}^{obj}$ to generate colour context region $\mathbf{CC}_L$\;
  4. Crop image $\mathbf{I}_L$ at $x, y$ with mask $\mathbf{S}^{obj}$ to generate colour synthesis region $\mathbf{CS}_L$\;
  5. Generate edge map $\mathbf{E}_L = \operatorname{Canny}(\mathbf{CC}_L+\mathbf{CS}_L)$\; 
  6. Estimate depth map $\mathbf{D}_L = \operatorname{PSMNet}(\mathbf{I}_L,\mathbf{I}_R)$\;
  7. Crop image $\mathbf{D}_L$ at $x, y$ with mask $\mathbf{S}^{obj}$ to generate depth synthesis region $\mathbf{DS}_L$\;
  8. $disp = \max(\mathbf{DS}_L)$\;
  9. Reproject $\mathbf{C}^{obj}$ using $disp$ value onto $\mathbf{I}_R$
  and crop to generate stereo colour context region $\mathbf{CC}_R$\;
 }
 \caption{Stereo-aware training set generator}
 \label{alg:data-generation}

\end{algorithm*}
Several methods~\cite{Nazeri2019,Ren2019,song2018} have shown that structure-guided inpainting performs better at reconstructing high frequency information accurately. Since image structure is well-represented in its edge mask, superior results can be obtained by conditioning an image inpainting network on edges in the missing regions. For this reason, we feed the edge maps generated using Canny edge detector~\cite{canny1986} along with the colour information, as a bias to our network, following a similar process to Nazeri \etal~\cite{Nazeri2019}. At test time, we estimate the edges using a pre-trained \textit{EdgeConnect}~\cite{Nazeri2019} model.

\subsection{Stereo consistency Loss}\label{subsec:disp_loss}
Inspired by the work of Chen \etal~\cite{chen2019}, we propose a local consistency loss which measures the consistency between the inpainted area in one view, and the ground truth in the other view. In this way, we encourage the system to use the stereo context; inpainting not just any perceptually acceptable background, but specifically the one consistent with any partial observations.
The loss is illustrated in Fig.~\ref{fig:model_overview}.

We compare a patch $P\left(i\right)$ around every pixel $i$ in the inpainted area $\mathbf{S} \odot \mathbf{I}$ against a patch centred on the corresponding pixel on the other view. $\mathbf{S}$ is the binary mask indicating the synthesis region, $\mathbf{I}$ is the inpainted image, and $\odot$ denotes the Hadamard product.
\begin{align}
    \mathcal{L}_{disp}&=\frac{1}{\left| \mathbf{S}\right|}\sum_{i \in \mathbf{S} \odot \mathbf{I}}\overleftarrow{cost}\left(i\right), \\
    \overleftarrow{cost}\left(i\right)&=1-\Phi\left( P\left(i\right), P\left(\overleftarrow{W}\left(i\right)\right)\right)
\end{align}
where $\overleftarrow{W}$ is the warping function corresponding to a change from source to target view, using the disparity estimated by \textit{PSMNet}~\cite{chang2018}. We use a Normalised Cross-Correlation (NCC) as our stereo matching cost ($\Phi$) which works well with back-propagation.
\begin{equation}
    \Phi(X,Y)=\dfrac{\left\|X \odot Y\right\|_{1,1}}
        {\left\|X\right\|_F\left\|Y\right\|_F}
\end{equation}
here $\left\|\cdot\right\|_{1,1}$ and $\left\| \cdot \right\|_{F}$ are the 1-entrywise and Frobenious matrix norms respectively.

\subsection{Inpainting losses}

In addition to the disparity loss, other per-pixel similarity losses and losses based on deep features are used to enforce perceptually realistic results. First, two per-pixel reconstruction losses are defined over the synthesis and context regions, these losses help guiding the inpainting of the missing areas, as well as making sure that context and synthesis areas are recovered consistently and with smooth boundaries.
\begin{align}
\mathcal{L}_{synthesis} &= \frac{1}{N_{\mathbf{I}_{gt}}} \left\|\mathbf{S} \odot \left(\mathbf{I} - \mathbf{I}_{gt}\right)\right\|_1, \\
\mathcal{L}_{context} &= \frac{1}{N_{\mathbf{I}_{gt}}} \left\|\mathbf{C} \odot \left(\mathbf{I} - \mathbf{I}_{gt}\right)\right\|_1
\end{align}
where $\mathbf{S}$ and $\mathbf{C}$ are the binary masks indicating synthesis and context regions respectively, $N_{\mathbf{I}_{gt}}$ is the total number of pixels, $\mathbf{I}$ is the inpainted result, and $\mathbf{I}_{gt}$ is the ground truth image.
In addition, we include two deep feature losses from Johnson \etal~\cite{johnson2016}, based on VGG-16~\cite{simonyan2015} embeddings, that measure high-level perceptual and semantic differences. Firstly
\begin{align}
\mathcal{L}_{perceptual} = \sum_{p=0}^{P-1}\frac{\left\| \Psi_p\left(\mathbf{I}\right)-\Psi_p\left(\mathbf{I}_{gt}\right)\right\|_1}{N_{\Psi_p}}
\end{align}
where, $\Psi_p(\cdot)$ is the output of the \textit{p}'th layer from VGG-16 \cite{simonyan2015}, and $N_{\Psi_p}$ is the total number of elements in the layer.
Secondly, the style loss is defined as,
\begin{align}
\mathcal{L}_{style} = \sum_{p=0}^{P-1}\frac{1}{C_pC_p}\left\|K_p \left[\left(\Psi^{\mathbf{I}}_p\right)^{\intercal}\Psi^{\mathbf{I}}_p-\left(\Psi^{\mathbf{I}_{gt}}_p\right)^{\intercal}\Psi^{\mathbf{I}_{gt}}_p\right]\right\|_1
\end{align}
where $K_p=\frac{1}{C_pH_pW_p}$ is a normalisation factor, and $C_p, H_p, W_p$ are the number of channels, height, and width of the output $\Psi_p(\cdot)$.

These perceptual losses encourage the network to create images with similar content and similar feature representations. The style loss ensures that the style of the output images resemble the input in colour, textures, etc. Finally, a total variation loss is used as a smooth regularization.
\begin{align}
\mathcal{L}_{tv} &= \sum_{\left(i,j\right)\in \mathbf{S}}\frac{\left\| \mathbf{I}(i,j+1)-\mathbf{I}(i,j)\right\|_1}{N_{\mathbf{I}_{gt}}} \\
&+ \sum_{\left(i,j\right)\in \mathbf{S}}\frac{\left\| \mathbf{I}(i+1,j)-\mathbf{I}(i,j)\right\|_1}{N_{\mathbf{I}_{gt}}}
\end{align}
where the $\mathbf{S}$ denotes the pixels in the synthesis region.

\begin{figure*}[htbp]
	\begin{center}
        \includegraphics[width=\linewidth]{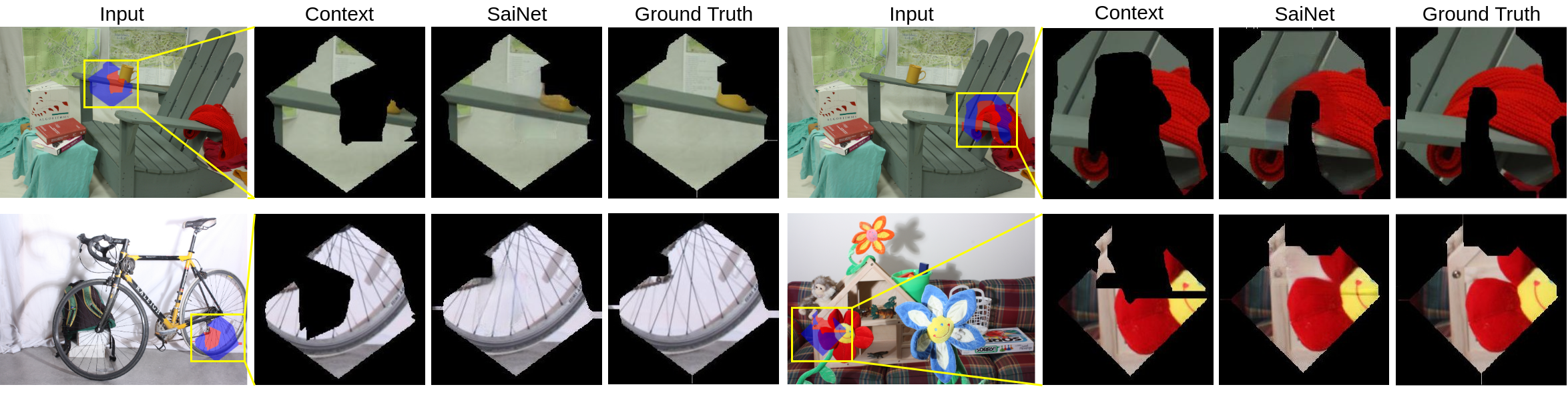}
    \end{center}
	\caption{\textbf{Real dataset evaluation} over Middlebury~\cite{jensen2014} using Canny edge detector. The zoomed-in crop of yellow area is visualised as ``Ground Truth''.}
	\label{fig:qualitative_evaluation_middlebury}
\end{figure*}

Similar to Liu \etal~\cite{liu2018}, we use the following weights to combine all these losses to yield the final training objective: $\lambda_{synthesis}=6$, $\lambda_{context}=1$, $\lambda_{perceptual}=0.05$, $\lambda_{style}=120$, $\lambda_{tv}=0.1$, $\lambda_{disparity}=0.1$. The same parameters are used for all evaluations.

\subsection{Datasets}

Good quality, natural stereo datasets are very hard to come by. This is a problem for training deep neural networks, which usually require a high number of images to extract meaningful statistical information. Our approach to data collection intrinsically performs data augmentation, as the random sampling of context-synthesis areas makes it possible to use different samplings of the same image without overfitting.

For training we have used three different datasets: SceneFlow~\cite{mayer2016}: \textit{FlyingThings3D}, \textit{Driving}, and \textit{Middlebury}~\cite{scharstein2014}. \textit{FlyingThings3D} consists of 21,818 frames from 2,247 scenes, containing everyday objects flying around in a randomised way. This is ideal for training CNNs due to the large amount of data and variety of objects. \textit{Driving} is a more naturalistic-looking dynamic street scene resembling the \textit{KITTI} dataset~\cite{geiger2012}. It contains 4400 images from one scene. On the other hand, the \textit{Middlebury} dataset consists of only 33 pairs of stereo images of natural scenes. Even though this dataset is not big enough to train a Deep Learning model, we are able to perform transfer learning and generate pleasant results over real world data (See Fig.~\ref{fig:qualitative_evaluation_middlebury}).

These datasets contain ground truth disparity maps, but for our model we have included a disparity estimation step using \textit{PSMNet} so we don't rely on existing ground truth data. This makes it fairer to compare to other models that use a similar approach, as well as being more relevant to our application to media production, where we may have several views from the same scene, but no depth information.

\begin{figure*}[hbtp]
	\begin{center}
        \includegraphics[width=\linewidth]{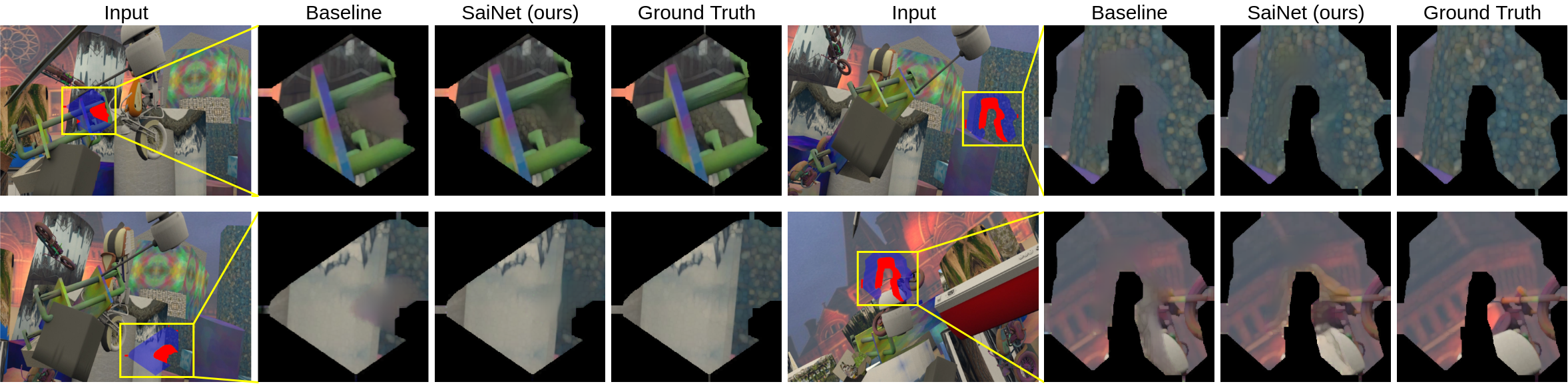}
    \end{center}
	
	\caption{\textbf{Qualitative inpainting results} for FlyingThings3D. Baseline is Shi \etal \cite{Shih2020}. The zoomed-in crop of the yellow area is visualised in the ``Ground Truth'' column.}
	\label{fig:qualitative_evaluation}
\end{figure*}

\begin{table*}[hbtp]
    \captionsetup{skip=3pt,width=.7\textwidth}
	\caption{\textbf{Quantitative results.} Image quality \& stereo consistency of different models. \textbf{Bold} is best. $\star$ values are from their paper.}
	\centering
	\begin{tabular}{ llcccc } 
	\toprule
	 Dataset                          & Model & PSNR$\uparrow$ & SSIM$\uparrow$ & LPIPS$\downarrow$ & DispE (\%)$\downarrow$ \\
	\midrule
	\multirow{2}{*}{FlyingThings3D}
	& Shih \etal~\cite{Shih2020} & 28.32 & 0.8589  & 0.0707             & 7.96 \\
	& Ours                       & \textbf{30.50} & \textbf{0.8643} & \textbf{0.0556} & \textbf{7.67} \\ 
	\midrule
	\multirow{3}{*}{Driving}
	& Shih \etal~\cite{Shih2020}         & 30.46  & 0.969 & 0.1141 & 9.94 \\ 
	& Chen \etal~\cite{chen2019}$^\star$ & 22.38  & 0.959 & - & 7.79   \\ 
    & Ma \etal~\cite{ma2020}$^\star$     & 23.20  & 0.964 & - & \textbf{4.72} \\ 
	& Ours                       & \textbf{34.94} & \textbf{0.977} & \textbf{0.0628} & 8.01 \\
	\bottomrule
	\end{tabular}
	\label{table:quantitative_evaluation}
\end{table*}

\begin{table*}[htpb]
    \captionsetup{skip=3pt,width=.7\textwidth}
	\caption{\textbf{Ablation study.} Compare the accuracy of different stages of the model over all regions. `Baseline' is the monocular inpainting model, `Stereo' is the model + stereo context, `Disp' is the model + disparity loss, and `Full' if the model with both stereo context and disparity loss. \textbf{Bold} is best result. \textcolor{blue}{Blue} are results in synthesis regions only.}
	\centering
	\begin{tabular}{ lcccc }
	\toprule
     Model             & PSNR$\uparrow$ & SSIM$\uparrow$ & LPIPS$\downarrow$ & DispE (\%)$\downarrow$ \\
	\midrule
	Baseline      & 28.32 (\textcolor{blue}{22.26}) & 0.8589 (\textcolor{blue}{0.5619}) & 0.0707 (\textcolor{blue}{0.0676}) & 7.96 \\
	Ours (Stereo) & 29.41 (\textcolor{blue}{23.61}) & 0.8604 (\textcolor{blue}{0.5597}) & 0.0625 (\textcolor{blue}{0.0582}) & 7.68 \\
	Ours (Disp)   & 29.79 (\textcolor{blue}{24.00}) & 0.8619 (\textcolor{blue}{0.5684}) & 0.0570 (\textcolor{blue}{0.0569}) & 7.71 \\
	Ours (Full)   & \textbf{30.50} (\textcolor{blue}{\textbf{24.70}}) & \textbf{0.8643} (\textcolor{blue}{\textbf{0.5771}}) & \textbf{0.0556} (\textcolor{blue}{\textbf{0.0539}}) & \textbf{7.67} \\
	\bottomrule
	\end{tabular}
	\label{table:ablation_study}
\end{table*}

\subsection{Experiment setup}\label{subsec:experiment_setup}

The network is trained using a batch size of 8 and $256\times256$ images. The model is optimised using Adam optimiser~\cite{kingma2014} and a learning rate of 0.1. A model has been trained for each different dataset. As \textit{FlyingThings3D} is 3 to 5 times bigger than the other datasets, a transfer learning approach has been followed where the model is trained on \textit{FlyingThings3D} first and then fine-tuned over \textit{Driving}, and \textit{Middlebury}.

For fair comparison to the results of Chen \etal~\cite{chen2019} and Ma \etal~\cite{ma2020}, we have trained our \textit{Driving} model using $128\times128$ square context masks and $64\times64$ centred synthesis masks. Our baseline model is Shih \etal~\cite{Shih2020} 3D photography colour inpainting network, which has been trained in the same fashion as our model, and conditioned over depth edges instead of colour as per their original pipeline.

For training, we generate edge maps using Canny~\cite{canny1986} edge detector following \textit{EdgeConnect}~\cite{Nazeri2019} approach. At test time, we apply pre-trained \textit{EdgeConnect} models to generate the synthesis area edges, using the pre-trained model over Places2~\cite{zhou2018b} for our \textit{FlyingThings3D}, and \textit{Middlebury}, and a pre-trained model over Paris StreetView~\cite{doersch2012} for our \textit{Driving}.

\section{Results and Discussion}
In this section we show different evaluations and comparatives that demonstrate the value of our work. We train our model on three different datasets as explained in Section \ref{subsec:experiment_setup}, and we compare its accuracy and consistency against state-of-the-art methods. We also perform an ablation study to evidence the  benefits of the different contributions of our model.

\subsection{Evaluation of Accuracy}
There is no perfect numerical metric to evaluate image inpainting outputs given the variety of possible valid results. For the purpose of quantifying how well our model performs, we make use of several popular metrics that measure different characteristics of an image. To measure image quality, we use Peak Signal-To-Noise Ratio (PSNR)~\cite{huynh-thu2008} and Structural SIMilarity (SSIM)~\cite{wang2004} index. PSNR shows the overall pixel consistency, while SSIM measures the coherence of local structures. These metrics assume pixel-wise independence, which may assign favourable scores to perceptually inaccurate results. For this reason, we also include the use of a Learned Perceptual Image Patch Similarity (LPIPS)~\cite{zhang2018a} metric, which aims to capture human perception using deep features.

The stereo consistency is quantified using the disparity error metric from \cite{ma2020}, which counts the erroneous pixels of the \textit{PSMNet} estimated disparity map of the inpainted image, compared against the ground truth\footnote{The definition of \cite{ma2020} has a typo where the absolute error $\left|d^{i}_{est}-d^{i}_{gt}\right|$ is replaced by $d^{i}_{est}$.}. Given the inpainted image $\mathbf{I}$, for every pixel $i$ we consider its estimated disparity $d^{i}_{est}$ to be erroneous iff the absolute error against the equivalent pixel in the ground truth disparity image $d^{i}_{gt}$ is greater than $p_1$ and its relative error greater than $p_2$ (we use $p_1=3$ and $p_2=0.05$). This is described in equation \ref{eq:disp_e}, where $N$ is the total number of pixels, and $[\;]$ is the Iverson bracket.
\begin{align}
    \label{eq:disp_e}
    \mathit{DispE} = \frac{1}{N}\sum_{i \in \mathbf{I}} \Biggl[\Bigl(\left|d^{i}_{est}-d^{i}_{gt}\right| &> p_1\Bigr)\\
    \And \Biggl(\frac{\left|d^{i}_{est}-d^{i}_{gt}\right|}{d^{i}_{gt}} &> p_2\Biggr) \Biggr]
\end{align}

\subsection{Inpainting comparison}
We perform a quantitative comparison of our inpainting model against other state-of-the-art methods~\cite{Shih2020,chen2019,ma2020}, following the experiment setup described in Section \ref{subsec:experiment_setup}. Results can be seen in Table \ref{table:quantitative_evaluation}.

We can observe our model performs better across all metrics compared with the baseline model of Shih \etal~\cite{Shih2020}. Our model also performs competitively against other stereo inpainting models~\cite{chen2019,ma2020}, showing a superior image inpainting quality with an improvement on PSNR values of 50\%, and some improvement to SSIM. Due to the nature of our mask generation process, our stereo context information is quite narrow, limiting the visible area that our network can learn from. Despite this, our model accomplishes similar results to the stereo consistency of Chen \etal~\cite{chen2019}.
The image quality of the inpainting is superior on the Driving dataset, which was trained using square centred masks to match the experimental setup of~\cite{chen2019,ma2020}. Meanwhile, the object-like occlusion masks used on the FlyingThings3D dataset, which are not fully bounded, are much more challenging.

A qualitative example is shown in Figure \ref{fig:qualitative_evaluation}\footnote{For further results and analysis we refer the readers to the supplementary material}. Despite having access to depth edges, Shi \etal struggles to produce sharp object boundaries in the inpainted region. Meanwhile SaiNet is able to use stereo context to inpaint sharp boundaries using colour edge information. This is evidenced by Shih \etal success recovering the green bar in the first example, but failing on the colour edge of the second example. However, as shown in the 4th example, our technique still struggles to inpaint especially intricate structures which are not visible through stereo context. Nevertheless it produces sharper and more visually pleasing results.

\subsection{Ablation Study}
In the interest of proving the contribution of every stage to the accuracy of the model, we have studied its performance removing the key contributions. Results presented in Table \ref{table:ablation_study} show that every part of the model performs better than the baseline, with the combination of all modules having the best performance across all metrics. It is interesting to note that the use of a disparity loss provides the largest individual benefit in terms of stereo consistency.

\section{Conclusion}
We introduced a new stereo-aware learned inpainting model that enforces stereo consistency on its output, trained in a self-supervision fashion over geometrically meaningful masks representing object occlusions. This technique improved over state-of-the-art models by up to 50\% PSNR, and we demonstrated its performance over several diverse datasets. As future work, it would be helpful to explore how we could extend similar techniques to cope with the challenges that wide-baseline non-parallel cameras would provide.

\paragraph{Acknowledgements.}This work was partially supported by the British Broadcasting Corporation (BBC) and the Engineering and Physical Sciences Research Council's (EPSRC) industrial CASE project ``Generating virtual camera views with generative networks'' (voucher number 19000033).

{\small
\bibliographystyle{ieee_fullname}
\bibliography{vmgbib.bib}
}

\end{document}